# SYNTAGMA: a Linguistic Approach to Parsing

Daniel Christen
(www.lector.ch)

**Abstract**

SYNTAGMA is a rule-based parsing system, structured on two levels: a general parsing engine and a language specific grammar. The parsing engine is a language independent program, while grammar and language specific rules and resources are given as text files, consisting in a list of constituent structures and a lexical database with word sense related features and constraints. Since its theoretical background is principally Tesnière's *Éléments de syntaxe*, SYNTAGMA's grammar emphasizes the role of argument structure (*valency*) in constraint satisfaction, and allows also horizontal bounds, for instance treating coordination. Notions such as Pro, traces, empty categories are derived from Generative Grammar and some solutions are close to Government&Binding Theory, although they are the result of an autonomous research. These properties allow SYNTAGMA to manage complex syntactic configurations and well known weak points in parsing engineering. An important resource is the semantic network, which is used in disambiguation tasks. Parsing process follows a bottom-up, rule driven strategy. Its behavior can be controlled and fine-tuned.

## 1. Introduction

The current panorama of parsing techniques shows some very elegant mathematical models and algorithms, and their implementation gives in many cases interesting results regarding correctness of syntactic and semantic interpretation of the input[1]. Nonetheless these are still incomplete (in spite of the continuous growth of tree banks) and need post-editing and special procedures in order to neutralize artificial asymmetries on the different levels of representation. Most of these parsers are data driven; therefore their behavior can hardly be controlled, and they fail when statistical weights outperform special rules which should be applied. In many cases theoretical discussions and parsing problems seem not to be a consequence of the structural complexity of language, but a product of the adopted formalism (both in parsing and in representation). Discussion, for example, about crossing arcs (which some consider a marginal phenomenon) in dependency parsing, is a consequence of the choice to treat coordination as a kind of dependency relation, instead of considering it a symmetrical one. With an adequate parsing technique, the syntactic representation matches the semantic representation without the need of *swap* or other kind of techniques which are often seen in Natural Language Processing[2].
In this paper I will describe SYNTAGMA, a parsing system provided with an autonomous PoS-tagger, which moves from a linguistic frame instead of a statistical one. The basic assumption is: you cannot reach good results in syntactic parsing without pragmatic, textual and semantic information; and you cannot not achieve semantic tasks, like semantic role assignment or WSD, without adequate syntactic information[3]. The theoretical background is given by Tesnière's work[4], but I use categories, concepts and formalisms that also come from other syntactic theories (Government & Binding for example), and from semantic research.

---

1    J. Nivre (2005) provides for a wide-ranging review of the state of the art in dependency parsing.
2    The author of this article contributed to the development of a dependency parser, a hybrid rule-based and data driven parser (M. Grella, M. Nicola, D. Christen, 2011), which achieved best score at Evalita 2011 Dependency Parsing Task. Strengths and weaknesses of dependency parsing appeared clearly, despite these results.

In the next sections I will discuss SYNTAGMA's grammar-levels, syntactic constituents, relation types and constraints filter. I will also discuss the solution I propose for some typical parsing problems, such as attachment ambiguities, coordination, gaps[5].

## 2. SYNTAGMA's grammar

### 2.1 Two levels of syntactic representation

An input text or sentence in natural language is a linear sequence of words. Linearity (time-linearity in spoken, space-linearity in written language) causes an asymmetry between the expression level and a structural-semantic level of interpretation and comprehension. The most important consequences of this asymmetry are: a) relations are not necessarily between adjacent elements: linear contiguity is not always a relation mark; b) there is a lot of information that may be "deleted" on the expression level, but which is implied somehow and which a human being normally fills with information that comes from other parts of the sentence or from the co-text, or even from the situational context. "To understand a sentence means to translate the linear order into a structural order"[6].

In SYNTAGMA's grammar there are two levels of syntactic representation. The first, which I call constituent level, assures the link between the surface-expression level and its structural and functional interpretation; a list of patterns for each constituent type, describes its possible linear configurations on the expression level; those configurations are related to the constituent's dependency and functional structure. The second, called representational level, is a conventional formalism for representing the interpretation of the given surface structure.

I) The *constituent level* describes, for each constituent contemplated in this grammar, the sub-constituents which belong to its structure and their linear order. Syntactic constituents are described as patterns, with a given sub-constituent order, their dependency relations and their syntactic functions (section 2.2.8). Constituent pattern structure is parallel to the linear level: the constituent sequence follows exactly the surface linear order. Since dependency relations and syntactic functions (i.e. the structural and functional interpretation of the elements of a pattern) are directly defined on the constituent level, there is no need for "transformations" or "movements". Constituent level plays a role only during the parsing process: it guides the process by tagging a sequence of words which satisfy some given conditions and constraints, and by assigning to each constituent its syntactic function and its place in the dependency tree.

II) The *representational level* (the output of the parser) is given in the form of an indexed list of words, provided with their grammatical features, their function and their syntactical relationship to other words through an index. This representation may be translated in dependency graphs but also in non-planar graphs. Words are nodes and syntactic and semantic relations are the arcs between nodes. Post-parsing harmonization between the syntactic relations frame (supposed to be planar) and the semantic relations frame (supposed to be non-planar) is not needed, since these frames are directly related on the level of constituent description. Thus, a problem which is often considered as a structural one, has here a purely representational *status*. And since the argument frames related to the lexical entries (in the case of verbs, first of all, but also nouns and adjectives may have an argument frame) is semantically tagged, SYNTAGMA's output can be easily translated into a semantic tree or graph.

---

3   The reader will find the sames conclusions in D. Gildea and Martha Palmer, *The necessity of Parsing for Predicate Argument Recognition*, 2002; and V. Punyakonok, D. Roth, W. Yih, *The necessity of Syntactic Parsing for Semantic Role Labeling*, 2005.
4   L. Tesnière, *Éléments de syntaxe structurale*, Paris 1959. Curiously, Lucien Tesnière is one of the most referred authors in Dependency Parsing literature, although hes intuitions and notions are one little applied in practice.  The view on Tesnière's theory from the perspective of Cognitive Grammar is more congruent: Langacker 1994.
5   The technical description of SYNTAGMA's architecture, its resources and tools are available at http://www.lector.ch.
6   "Toute la syntaxe structurale repose sur les rapports qui existent antre l'ordre structural et l'ordre linéaire. Construire ou établir le stemma d'une phrase, c'est en transformer l'ordre linéaire en ordre structural" (Tesnière, 1959, chap. 6).

**2.2. Constituents**

2.2.1 Constituents are *projections*[7] of their lexical heads. This means that word argument structure (i.e. *valency*) plays a fundamental role in this grammar[8].

2.2.2 On the representational level, only semantical "full" words are allowed to behave as heads, that is: verbs, nouns, adjectives, adverbs and, on the intermediate level, determiners[9].

2.2.3 Constituent tags are related to their heads: V > C; N > NP; Adj > AdjP; Adv > AdvP. Therefore, the syntactic output gives only these four types of fundamental constituents[10]. There are as well subtypes, which have mostly only an empirical status or a practical utility, such as Coord, an empirical constituent which handles coordinated structures; and some intermediate constituents, such as Det (determiner), Sep (separator). Also a PP constituent may be introduced for practical purposes, but it will not have theoretical and structural *status*. Those constituents are ephemeral: they may exist in a given intermediate state of the parsing process but are soon absorbed by one of the four main constituent tags. Secondary constituents are pronouns, which are head of a NP, and relative pronouns, which are treated here as simply connectives. The abstract, i.e. non-lexical, S (Sentence) frame is the top element, containing information related to utterance and sentence type (assertive, interrogative, etc.) and the cues to its textual frame.[11]
A larger variety of distinctions is needed on the level of terminal constituents, specially in the closed classes, in order to allow a very fine-tuned description of constituent configurations (patterns): for instance between the different kind of words that can build a Det constituent ("all this...", "a certain...", "both these two..."). This classification takes place in the lexical database of the system[12].

2.2.4 From principles 2.2.1 and 2.2.2 follows an important distinction between:

*a) arguments*, that are constituents required by the argument structure (*valency*) of a given word and to

---

7   I use the notion *projection* in the meaning it have in Generative Grammar, e.g., the features of words (like the argument structure or *valency*), which model the syntactic configuration they generate. This notion has nothing to do with the sense the term *projectivity* has in dependency parsing literature.
8   The importance of the concept of valency in syntax was first emphatized by Tesnière (1959). Since then it has played a central role in grammar, for instance in Fillmore's Case Grammar (1968), in the different Dependency Grammar frameworks (since Hay 1965), in Cognitive Grammar, and finally also in Chomsky's generative framework, since the Government & Binding Theory (Chomsky, *Some Concepts and Consequences of the Theory of Government and Binding*, MIT Press, Cambridge Mass., 1982).
9   Those are the four "fundamental categories": Tesnière, p. 114. Therefore, a Prep can't be a head, and there is no PP constituent in this grammar.
10  According to Tesnière (1959, chapt. 114).
11  Analogous to the "technical root node" in the tectogrammatical tree of the Prague Dependency Treebank.
12  The fine classification of Det terminal elements, that is used in SYNTAGMA for Italian, is derived from G. Genot, *Grammaire de l'Italien,* Paris 1973.

which this lexical head can assign a syntactic function (subject, object...) and a semantic role (agent, patient...). We also call them *complements.* Arguments are necessary elements of the structural frame of a lexical entry (verbs and also of many nouns and adjectives), because they have a meaning-inherent function. They can, in some cases, be left unexpressed: thus, in SYNTAGMA's syntactic database, a dedicated slot provides for that information: optionality of some arguments may be true or false.

The minimal composition of a constituent is given by its head and its arguments: it constitutes the *nucleus* (*nuclear phrase or structure*).

b) *adjuncts (expansions)*, which are optional modifiers of a nuclear structure. They are not contemplated in the argument structure of the lexical head, and their function is to add some additional information to the nuclear constituent. This information may have a restrictive[13] or a circumstantial/adverbial funtion[14]. Expansions do not play a meaning-inherent role, thus their *status* is not the same as the one of arguments.

2.2.5 A nuclear phrase may be *expanded* by optional modifiers (*expansions/adjuncts*). Their head depends of the head of the *nucleus*. From a representational point of view, in the dependency graph, there is no distinction between arguments and expansions. The difference is structural and belongs to the parsing process, while argument structure satisfaction is a discriminating factor between correct and incorrect interpretations of the input.

2.2.6 The distinction between nuclear elements (that are needed by the argument structure of the lexical head) and optional expansions (adjuncts) plays instead a crucial role on the level of semantic interpretation and selection of the parsing result, since each meaning of the lexical head has its specific argument structure. Semantics and syntax are therefore strictly related.

In SYNTAGMAS' lexical database the syntactic features, i.e. argument structure and all kinds of restrictions over arguments (such as morphological, positional and semantic restrictions) are meaning-related. This allows that lexical head's meanings may be selected considering their correspondence with the parsing result.

2.2.7 Constituent linear order, which has important cross-linguistical variations, is described in an independent file, simply a text file, a resource of the system. This allows a language focused *parametrisation* of this linguistic dimension, and makes SYNTAGMA a potentially multi-language system.

As mentioned, constituents are described in their linear configuration as patterns, with a given subconstituent order, their dependency relations, their syntactic functions and constraints. Therefore, for each language, its specific syntactic structures and features can be described independently, and do not affect the core of the parsing system.

From a theoretical point of view, this gives a two level grammar. The first one, the *general grammar*, describes language features which are in a way supposed to be universals: recursion, projection of argument structure, distinction between arguments and adjuncts, general types of relations (symmetrical, asymmetrical), possible strategies for case assignment (lexicalisation, grammaticalisation, positionality, prosody). The second level is *language-focused grammar* and allows different *parametrisation* of morphological proprieties, of constituent patterns, of relation marking and of semantic case assignment.

---

13  One may represent the restrictive function of modifiers (adjectives, specifiers or relative clauses) as an operation on sets, since a noun designs a class of elements, and modifiers add a characteristic which generates a subset resulting from the intersection between the set of elements designed by the noun and the set of things owning that propriety. The reader is referred to G. Chierchia and S. McConnel-Ginet, *Meaning and Grammar,* 1990. This set theory approach has been didactically transposed in D. Christen, *Grammatica e matematica non fanno solo rima*, "Italiano&Oltre" (ed. R. Simone), 3-4, 2000, pp. 152-159.

14  This optional constituents are also called "circumstantials" and "adverbials" for example in Renzi et. al. (1988): they relate the nuclear content to temporal, spatial, manner, quality or other kind of additional information.

2.2.8 A constituent pattern frame is a structure which contains four sets of data: the linear sequence of sub-constituents, the internal dependency relations, the syntactical functions, and the constraints.
The two simple examples below, show a variety of NP pattern and a variety of finite clause pattern:

```
NP    { (Det, Adj, N),   (3, 3, 0),   (det,  adj, head),   (agr(num, gen), agr(num, gen), nil) }
         ↓                ↓            ↓                     ↓
         linear seq.      depend.      functions             constraints
```

The frame here above provides a description for a subtype of NP, which consists in the linear sequence of a Determiner, an Adjective and a Noun (which are syntactical categories, some may be terminal some intermediate categories). The dependency tree is, at this level, position referred: the first element and the second element have as their head the third one, which has no governor yet. Functions are assigned in the third set: to each position in the sequence corresponds a specific function. The fourth set contains the constraints: which in our example is the agreement that the head the dependent nodes (Det and Adj) must satisfy (in Italian, for instance, but not in English).

```
C     { (NP, V, NP), (2, 0, 2), (subj, v, obj),  (agr(pers, num), nil, nil) }
```

This basic clause pattern shows a linear sequence done by an NP, followed by a Verb and another NP in the third position. Their dependency relations are described in the second set: both, the first and the third terms have the second (the Verb) as their head. Syntactic functions are assigned in the third set. In the fourth set, corresponding to the subject function, the subj-verb agreement is formalized; the other two positions are not submitted to constraints.

Since the SYNTAGMA parser works recursively bottom up, and generates, cycle after cycle, new constituents by combining those generated during the precedent cycles, constituents may be described following the classical derivation formalism. And therefore it is possible to easily describe all kinds of variants of complex structures using few symbols. Constraints may be "relaxed" in a controlled way, in order to open the system to more informal utterances or even to accept norm deviances which are typical of some contexts (spoken language, social network communication etc.).

**2.3  Relations: some fundamental notions**

2.3.1 There are two kinds of structural relations between syntactical constituents: asymmetrical relations, called *dependency relations*, and symmetrical relations like *coordination*. Crossing arcs are allowed, not only with respect to phenomena which Generative Grammars relate to *movements*, *gaps* etc., but also because they are the necessary consequence of two coordinated heads having the same dependent ("James builds and repairs computers", "Good drinks and food").

2.3.2 A fundamental assumption is that relations may be expressed in three ways in (written) language: a) by *lexicalisation*, b) by *morphology*, c) by *linear-position*[15]. In spoken language you have also prosodic

---

15  Tesnière (*Élements..*, chapt. 111) notices how genitive is treated in Latin ("liber Petri"), French ("le livre de Pierre") and English ("Peter's book") grammars: by classifying and describing it in different ways (one as a morphological phenomena, one as a PP, one a saxon-genitive), the fact that it concerns the same semantic relation takes second place. What we propose is a radical distinction between structural/semantic relations, and the strategies (lexicalisation, grammaticalisation and

(intonation) marks for relations, which become relevant first of all in detecting dislocations, topicalization and interrogative structures. We will consider here only the first three types of relation marks (employed in written text inputs).

a) *Lexical marks* for dependency are typically functional words ("empty words") like prepositions, subordinative conjunctions, and prepositional locutions. Their role is only to express a dependency relation, and in some cases, to express the semantic value of the relation ("because"=causal; "when"=temporal...). Thus there is no consistent reason for assuming a PP constituent instead of something which is nothing else but a NP whose dependency is lexicalized through a preposition; the more then prepositions do not have a univocal meaning[16]. In this grammar also relative "pronouns" are treated simply as connectives.

b) *Grammatical/morphological dependency marks* are used in subordinate clauses, often as a stylistic alternative to lexical relation marks (i.e. instead of a conjunction). The function of verbal moods like gerundive, participle, infinitive, besides giving some semantic information, are structural/syntactical, since they express a dependency relation. A lexical mark (preposition) may co-occur with a morphological mark (infinite verbal mood) in some subordinate clauses.

c) *Positional marks* may cross-linguistically substitute or alternate with inflectional marks, for example in subject/object case assignment in languages like English, Italian, French. Some languages like German use both, grammatical and positional case marking. Positional marks are also relevant in most of the dependency structures (det-noun, aux-verb, noun-adj, noun-apposition, noun-proper noun, noun-spec...).

2.3.3 Since languages may choose different strategies for expressing the same type of relation, this parameter must not be tuned in the general grammar, but on a language specific level, which means in the pattern description and in the syntactical features of the dictionary entries of the target language.

**2.4 Asymmetrical relations: dependency**

Following Tesnière and most dependency grammars, dependency is an asymmetric relation between the constituent head and one or more dependents, which are part of the constituent since they complete its sense (i.e. argomental constituents) or because they work as optional modifiers (adjuncts, circumstantials/adverbials). Both, head and dependent, are nodes of a relation (represented by an arc), which is designed to be vertical, so the whole configuration is given by a tree. Of course, this model is a conventional abstraction, but it has a theoretical, although not necessarily psychological, consistence, as it allows the explanation of many linguistic phenomena.
In the tradition of Dependency Grammars each dependent has only one head, and heads may have at least one ore more dependents. This leads to constraints and problems (like the discussion about *projectivity*) easily avoided by recognizing that natural languages are not limited by such a restriction. In the following

---

positionality) though which languages can express this relations. The reader will find a discussion of those notions in R.Simone, *Fondamenti di linguistica*, 1990, 35, 268, 333. This approach is applied in contrastive grammar by H. Glinz, *Grammatiken im Vergleich*, Tübingen 1994.

16 Languages may switch, from a diachronic point of view e, from morphological case-marking, like IOBJ dative in Latin, to a lexical one, and vice-versa. Case-marking uses cross-linguistically different strategies: morphological (and positional) mark for dative IOBJ in German, lexical mark for the same case in languages like Italian and French, only positional mark in English. And some expressions which should typically be expressed in form of a PP (like time expressions) are often a simply NP: "I want to visit Paris *one day*".

paragraphs it will be evident that multiple heads and crossing trees are very common phenomena, which appear in coordination, in anaphora relations and, obviously, in structures with dislocated elements (relative clauses, interrogative clauses and topicalisation). A same dependent constituent may even have different functions related to different heads. A sentence like: "John orders his dog to sit", implies that "dog" has two different heads: it is both IOBJ of "order" and SUBJ of "to sit" (although only in a deontic modality frame, from a logical perspective). If this structural (double headed dependency) is lacking, one could not generate the inferences needed to answer questions like: "Who does John order something to?" and "Who/what should sit?".

Theoretically, and also in practice, all four kinds of syntactic categories may be modifiers or adjuncts to each other. This allows *creativity* but also makes language a phenomenon which can be hardly reduced to coherent mathematical models, as every linguist knows[17].

### 2.5 Symmetrical relations: coordination

Coordination is a symmetrical (horizontal) relation[18]. Therefore a dependent constituent may have more then one head. Coordination is lexicalized ("and", "or"...). When lexical mark is lacking, pauses (comma separators in the written expression) may play the role both of separators between the coordinated elements, and of coordination marks. Coordinated elements have normally the same semantic content, which seems to be a condition for being on the same syntactical level[19]. Therefore it is possible to describe some formal criteria which enable disambiguation, when needed, of structures where coordination is involved.

### 2.6 Constraints, constraints filter and meaning selection

Constraints (restrictions, field: RSTR) which have to be satisfied by the dependent constituents are inscribed directly in the constituent pattern frame. These constraints may apply to all kinds of features: verbal mood, tense, morphological proprieties, and they can even restrict the lexical instance of a given element with a function like: *lex*="xxx". The following example illustrates a type of passive clause with overt agent ("This book has been written by Carver"):

| VP_pass: | Aux1 | Aux2 | V |
|---|---|---|---|
| ID: | 1 | 2 | 3 |
| RSTR: | (lex="have") | (lex="be",tmp=perf) | (mdv=part, tmp=past) |
| LEX: | "has" | "been" | "written" |

---

17  *Where do Grammars Stop?* (Labov 1972). The notion of *creativity* has a long tradition in philosophy of language and linguistic theory, from Humboldt to Saussure's *productivity*, and has been investigated from different perspectives, from Chomsky (1966), who formalizes Descartes concept of human creativity in terms of generative syntax, to Prieto (1967) and De Mauro (1971), who both inquire this notion in the framework of formal semantics. But eminent exponents of mathematical and formalistic approaches to language (from Ajdukiewicz and Tarski to Chomsky) always pointed out that the object of their investigations is *linguistic code, system, competence* (what Saussure calls *langue*), not linguistic *performance* and *utterances* (Saussure's *parole*). Languages have "more a problematic than a systemic nature" (De Mauro 1994). Therefore NLP systems should be so close as possible to the *plasticity* which natural languages show.

18  Tesnière calls it "junction": as it as not a qualitative but a quantitative phenomena, its node "has to be necessarily horizontal" in a graphic representation (1959, chap. 96).

19  Tesnière (1959, chap. 95) confirms this empirical intuition.

This subtype of a passive VP pattern contains first the verb "to have", present indicative, which is followed by a second auxiliary verb "to be", tense=perfect; the third element is the main verb, to which the past participle is imposed. The VP_pass constituent is the necessary constituent of the given subtype of passive clause:

| C_pass: | NP | VP_pass | Conn | NP |
|---|---|---|---|---|
| ID: | 1 | 2 | 3 | 4 |
| RSTR: | (agr) | - | (lex="by") | - |
| LEX | "this book" | "has been written" | "by" | "Carver" |

The first element in the pattern must be an NP which agrees with the second element, the VP; the second must be the VP_pass; next elements are a preposition (necessarily "by") and a constraints free NP.
Constraints may be formulated as sets of conditions, and provided with AND/OR operators for their fine tuning.
Since constituent pattern includes dependency and function assignment (2.2.8), the complete frame of this passive clause will be as follows:

| C_pass: | NP | VP_pass | Conn | NP |
|---|---|---|---|---|
| ID: | 1 | 2 | 3 | 4 |
| DEP: | 2 | 0 | 4 | 2 |
| FNCT: | obj | v_pass | conn | subj |
| LEX: | "this book" | "has been written" | "by" | "Carver" |

This frame may then be semantically completed by assigning the verb-meaning related attantial (i.e. thematic) roles: obj_pass = patient, subj_pass = agent.
In two steps SYNTAGMA does the job human beings probably do in only one step. First it checks if a given sequence of words matches with the structure of a constituent pattern. And after that it verifies if the given pattern type matches the argument structure and requested features of the head. This involves a selection also over the type of constituents which can be allowed and the connective (if requested) introducing the dependent constituent.
SYNTAGMA's lexical database is *meaning related* (2.2.6), thus grammatical features belong not to words but to meanings. A lexical head may have different meanings, each with a different argument frame and its special constraints: therefore the constraint filter checks if the given pattern type contains exactly the number and type of functions and the type of constituents that are requested by the argument structure of some word sense. The syntactic filtering mechanism has therefore also an important semantic function, because it selects from the set of meanings only those whose structural features match those of the actual input. In section 2.8.2 an example of this mechanism will be shown.

## 2.7 Co-reference

A third kind of relation we shall assume in this grammar is co-reference (e.g. anaphora), which links the structural level with the semantical level. It plays a central role in defining the antecedent of: a) pronouns b) possessive adjectives and pronouns c) empty categories (traces) d) non-overt subjects and objects [20] e) anaphora or cataphora phenomena on textual (inter-sentence) level, where referents are a part of a precedent/successive sentence or even the whole precedent/successive sentence itself. Constituents are provided with a dedicated co-reference slot and co-reference relations are formalized through a pointer to the related element.

---
20  L. Rizzi, *Null objects in Italian and the theory of pro*, "Linguistic Inquiry", 17, 1986, 501-58.

## 2.8 Complexity due to asymmetric nature of natural languages

As we have seen in section 2.1, natural languages show an fundamental asymmetry between the linear level of expression and the complex structure of logical and semantical relations that are represented. The inevitable flattening, on the linear level, of hierarchical (bi-dimensional) structures, added to the economy principle, characteristic of natural languages (leading to lacks of information on the linear level), are both the cause of ambiguities and interpretation difficulties concerning not only machines but people too. The different performance in this kind of problem solving goes back, in most cases, to a simply quantitative difference in general world knowledge, but also to something we call common sense, that is only apparently banal [21].

In the next paragraphs I want to briefly describe the resources, the data structures and the procedures used by the SYNTAGMA system to manage some of the most known difficulties in NLP.

### 2.8.1 Ambiguity in dependency relations

A well known problem in parsing tasks is the dependency ambiguity: for a given sequence A-B-C, both interpretations A(B(C)) and A((B)(C)) are structurally possible. There are two strategies for managing this problem, both employed by SYNTAGMA: a) using semantic information, which is the strategy naturally applied by human beings; b) using lists of word couplets or triplets and assigning a given weight to the relation which matches those data. This second strategy, practiced in statistical parsing, is used by SYNTAGMA only if the first kind of information (given in its semantic database) is lacking.

### 2.8.2 Gaps in dependency relations

In some languages and in some syntactic structures there are non-overt constituents: they are required by the argument structure of some lexical head, but they are not expressed on the surface-expression level. In Italian, for example, the non-overt subject is mostly a mark for anaphora and is therefore normal. In non-finite subordinate clauses gaps may interest one or more arguments of the non-finite verbal head. In the Generative Grammar framework those gaps are filled in the structural representation with the PRO constituent, which stays for the empty subject, and traces for objects. SYNTAGMA's grammar treats all kind of structural gaps as "empty categories" and fills them with traces which are related, so far as possible, to their antecedents by indexes. In some cases (for instance in passive clauses without an expressed agent) the trace-constituent is allowed to be left without a definite antecedent.

Traces are directly inscribed in the constituent pattern frames, which allows them to be assigned with their function and their constraints (for example agreement in number and genre with their antecedent in the case of traces in relative clauses). For non-finite clauses, the constituent frames contemplate variant patterns which contain all combinations of traces and overt constituents as needed by the argument frame of a verb. For example, the frame for a non-finite clause with a transitive verb as head shows the following variants:

---

21  This assumption has a central place in A.I. and cognitive science and was investigated among others by Miller, Shank, Minsky, Johnson-Laird; and discussed in the perspective of semiotics by Eco (1984) and Violi (1997). The reader will find some amusing examples in the first chapter of Pinker's *How the mind works*, 1997.

i) C  {(Tsubj, V, NP),   (1,0,1), (subj,v,obj), ()}      [*Paul wants*] to _ eat a hamburger
ii) C  {(Tsubj, V, Tobj), (1,0,1), (subj,v,obj), ()}     [*Which/the hamburger Paul wants*] _ to eat _

where *Tsubj* is the symbol for subj-trace and *Tobj* for object-trace. In variant i) only the subject lacks; in ii) both subject and object needed by the transitive verb are represented as traces. Following the Control feature of the head of the subordinate non-finite clause (in pour example "to want"), the algorithm for co-reference indexing makes *Tsubj* relate to "Paul" and *To* relate to "hamburger".

In the following example, the object lacks in the non-finite subordinate clause, and it is represented by a trace whose co-reference index points to the obj of the principal clause:

S:   "John bought a dog for Bill to give to Mary"
    = S(C(bought(John, dog, for(give(Bill, _ , to(Mary))))))
    = S(C(bought(John, dog$_i$, for(give(Bill, T$_i$, to(Mary))))))

| C(0)   | NP     | V        | NP      | C(1)      |
|--------|--------|----------|---------|-----------|
| ID:    | 1      | 2        | 3       | 4         |
| DEP:   | 2      | 0        | 2       | 2         |
| FNCT : | subj   | v        | obj     | adjunct   |
| RSTR:  | (agr)  | (agr)    | -       | (md=inf)  |
| COREF: | 0      | 0        | 9       | 0         |
| LEX:   | "John" | "bought" | "a dog" | -         |

| C(1)   | Conn       | NP    | VP        | NP         | Tobj      |
|--------|------------|-------|-----------|------------|-----------|
| ID:    | 5          | 6     | 7         | 8          | 9         |
| DEP:   | 7          | 7     | 2         | 7          | 7         |
| FNCT:  | conn       | subj  | v         | iobj       | obj       |
| RSTR:  | (lex="for")| (agr) | (md=inf)  | (conn="to")| (c=trace) |
| COREF: | 0          | 0     | 0         | 0          | 3         |
| LEX:   | "for"      | "Bill"| "to give" | "to Mary"  | -         |

In complex sentence structures, the syntactical function of the same element (i.e. its trace) may change from one level to the other, according to the argument frame of the respective verbal head. This phenomenon interests mostly modal verbs, causative verbs and verbs denoting a speech act expressing a request ("to ask x to do y"), which can take a subordinate non-finite clause as object. Control features, associated to the lexical entries, manage the co-reference assignment to traces with a different function in the lower clause. As example let us take the following (Italian) sentence:

iii) Paolo chiede a Giovanni di lasciargli prendere l'automobile ("Paul asks John to let him take the car")

Although SYNTAGMA's parsing mechanism works bottom-up, the operation modality of traces can be better explained looking at it from a top-down perspective. The main clause is parsed by selecting the following clause (C) constituent frame, since the elements of the given input match the constituent-sequence of this pattern. At the end of the first line of C-pattern, the constraint set which has been satisfied is shown; and below are dependency and function assignments:

| | C(0) | NP | V | NP | C(1) |
|---|---|---|---|---|---|
| ID: | | 1 | 2 | 3 | 4 |
| DEP: | | 2 | 0 | 2 | 2 |
| FNCT: | | subj | v | iobj | obj |
| RSTR: | | (agr) | (md=ind) | (conn="a") | (md=inf) |
| LEX: | | "Paolo" | "chiede" | "a Giovanni" | - |

The next step consists in evaluating if these patterns are consistent with respect to the argument frame of the verb "chiedere" ("to ask"). In SYNTAGMA's lexical database the different senses of this verb have the following argument structure and syntactic features:

mng 1.1 [ subj(cat(NP)),
      v(ctrl(false)),
      arg(cat(NP), opt(false
      prep.arg(conn("a"), cat(NP), opt(true))]

mng 1.2 [ subj(cat(NP)),
      v(ctrl(false)),
      arg(cat(C), mdv(cng), conn("che"), opt(false))
      prep.arg(conn("a"), cat(NP),opt(true))]

mng 1.3 [ subj(cat(NP)),
      v(ctrl(false)),
      arg(cat(C), mdv(OR(ind, cnd, cng)), conn("se"), opt(false))
      prep.arg(conn("a"), cat(NP), opt(true))]

mng 1.4 [ subj(cat(NP)),
      v(ctrl(false)),
      arg(cat(C), mdv(inf), conn("di"), opt(false)),
      prep.arg(conn("a"), cat(NP), opt(true))]

In our example, only the structure of meaning 1.4 matches the pattern C(0), and only this word sense reference will be registered in the set of meanings which is associated to this verb in the given sentence. Thus structure selection leads automatically to meaning selection. Note also that the meaning numerical index points to its semantic description in SYNTAGMA's semantic network.

Now the clause C(0) needs to be completed by the infinite-clause C(1). The control feature of the main verb "chiedere" ("to ask") has value=0, which means that its completive non-finite clause doesn't inherit its subject but its complements; thus the subject of the infinite clause C(1) has the co-reference index pointed to the indirect object of the principal clause:

| | C(1) | Prep | Tsubj | V | NP | C(2) |
|---|---|---|---|---|---|---|
| ID: | | 6 | 7 | 8 | 9 | 10 |
| DEP: | | 8 | 8 | 2 | 8 | 8 |
| FNCT: | | conn | subj | v | iobj | arg |
| RSTR: | | - | - | (mdv=inf) | (case=dative) | (mdv=inf) |
| COREF: | | 3 | (0,1) | 0 | 0 | 0 |
| LEX: | | "di" | - | "lasciare" | "gli" | - |

The argument structure of "lasciare", in its modal meaning, requires an indirect object to which subj function in the dependent clause may be assigned. Note, by the way, that "gli" has an ambiguous reference: it can as well refer to the subject of the sentence or to a third actor, mentioned in a previous sentence. But what interests here is its role in the underlying clause, since the causative verb "lasciare" ("to let") needs also a non-finite clause as complement. Therefore pattern C(2) is selected, as its restrictions (number and sequence of its constituents, verbal mood and null connective) match the given sub-sequence of the input.

"Lasciare" is a null-control verb, thus its subject is not inherited by the subordinate clause. Since subject lacks i C(2), the coreference assignment algorithm searches for a candidate in the upper structure C(1), and finds the constituent "gli" (9). Therefore the coreference slot of the subject trace of C(2) may be saturated as follows:

| C(2)   | Tsubj | V         | NP            |
|--------|-------|-----------|---------------|
| ID:    | 11    | 12        | 13            |
| DEP:   | 12    | 8         | 12            |
| FNCT:  | subj  | v         | arg           |
| RSTR:  | -     | (mdv=inf) | -             |
| COREF: | 9     | 0         | 0             |
| LEX:   | -     | "prendere"| "l'automobile"|

### 2.8.3 Intermediate traces

SYNTAGMA's grammar uses intermediate traces in the same way they are used in Government & Binding Theory[22], although I developed this solution autonomously to overcome some Italian specific structures like cliticization and argument raising in non-finite subordinate clauses.

These intermediate traces are also directly implemented on the level of constituent pattern description, providing for patterns in which some constituents are not arguments of the verbal head, but antecedents of empty categories in subordinate clauses, which may correspond to the linear result of raising procedures in a Generative Grammar framework.

In Italian, for example, also the object of a subordinate non-finite clause may be replaced by a clitic pronoun and raised on the level of the upper infinite clause ("Paolo chiede a Luigi di poterla prendere", "Paul asks John if he may take it") or even on the level of the main clause, typically when then main verb is a modal verb ("Paolo glielo vuole chiedere", "Paul wants to ask him for it").

Therefore the pattern list for subordinate clauses has to provide for this category of intermediate trace too. Coreference pointer assignment to the intermediate traces follows the same principles and mechanism shown in the previous section.

### 2.8.4  Ambiguity in coordinated structures

Criteria for resolving coordination ambiguities move from a semantical to a syntactical and morphological level, since the principle which allows words to be coordinated is a semantical one (coordination belongs to constituents which have a semantical similarity), which leads to category and/or morphological similarity (*congruence*). A verb, for example, may be modified by a sequence of constituents which express the manner (detected by a semantic tag), in the form of adverbs, but also NP (which refers to a manner and are generally introduced by a preposition) or subordinate clauses with gerundive inflection (which in given linear positions have a modal or temporal value). The structural level of a sequence of NPs, where all of them or only the first one may be preceded by a preposition ("[He graduated] in linguistics in Paris, [in] philosophy of language and [in] computer science in Rome"), can be reached only by applying semantical criteria (in our example: the distinction between disciplines and populated-places). It is often unclear if specifiers belong to both or only to one of the two coordinated words ("A thing made of wood and bones of some kind of animal"/"made of teeth and bones of some kind of animal"): also in these cases only the semantic similarity of the heads, for instance if both of them are meronyms of the specifier, can help in the disambiguation task[23]. Which means, in the second sentence, that the specifier must have two heads, if not yet on the level of syntactic structure, at least on the semantic representational level.

---

22   L. Haegeman, *Introduction to Governement & Binding Theory*, Oxford 1991, p. 463.

### 2.8.5 Gaps in coordinated structures

Coordinated structures, above all coordinated clauses, if you describe them as symmetrical relations, lead to crossing trees. From a structural point of view there are three fundamental types of trees: a) coordination of dependent constituents, for example the subject ("Paul and John love their parents") or the object ("Paul eats a hamburger and a salad"); b) coordination of the (verbal) head ("The children laugh and sing"), which can lead to a *plexus* configuration ("The machine generates and verifies its output") where both verbal heads share subjects and objects which are expressed only once, as dependents of the first, respectively of the second head; c) coordination interests verbal heads, but one of their dependents changes in the second coordinated clause while other dependents are shared by the heads ("Paul picks and Mary eats cherries", "Paolo raccoglie e Maria mangia le ciliegie", "Paul likes commendations and detests punishments, "Mario adora le lodi e detesta le punizioni", "Paul likes to be complimented and hates to be punished").

As regards these phenomena, SYNTAGMA's grammar does not follow Tesnière's description, which leads to very complicated *stemma*-configurations[24]. An alternative solution is to look at these structures as if they were affected by some kind of "deletion": like in other structural contexts, some information is semantically present while not expressed on the linear level. The solution I propose is close to the generative framework, by assuming the existence of gaps, which are replaced by indexed traces related to their overt antecedents.

The "deletion" mechanism in coordination follows the *given/new information* distinction, according to which a speaker deletes information that is given in the first member of the coordinate structure and will express, in the following member, only the *new* information, which is a *variant* of an argument or an adverbial constituent of the first member. Therefore the deleted element must correspond with the head of the first member at least, or with this head and one or more of its dependents.

Filling the gap, implies a procedure which has to a) detect the head of the first member of the coordinated structure; b) identify the syntactical function, with respect to this head, of the new elements introduced by the following member; c) identify, within the syntactical space of the first member, which dependents of the head are not replaced by new information; d) and finally fill the gap with a copy of the head and those last elements. This procedure needs syntactical information about the deleted lexical head, for instance its argument structure[25].

Deletion in coordinated structures can be predicted through SYNTAGMA's specific constituent frame descriptions, which are provided with traces (and then corresponding functional tags and features), which take the place of deleted constituents. Filling the gap consists in a saturation of the co-reference index of these traces, whose indexes relate them to their antecedents, i.e. the head of the first member and its redundant dependents.

---

23  Or also, of course, the shortcut used in statistically parsing, of binary head-dep word lists or triples head-prep-dep lists.
24  Tesnière 1959, chapters 103-107.
25  Some different approaches in Steedman (1996), Lombardo and Lesmo (1998), Dufour-Lussier, Guillaume and Perrier (2011).

**2.8.8. Controlled constraints relaxing and text-type specific system tuning**

Deviations from what is normally considered a linguistic rule in the area of grammar describing a certain language must be understood as a controlled relaxation of some well defined morphosyntactic and syntactic mechanisms. Involuntary errors or deviations and changes from the rule (for example in the area of informal communication) touch upon specific sectors of the code and may be precisely identified. In the area of informal communication for example in youngsters' jargon, if the speaker intends to be understood he uses a controlled release on certain linguistic ties (agreement for example), maintaining intact those necessary for comprehension of the message. Given that the deviations frome the rule are 'controlled' it is also possible to precisely identify which dimensions and grammatical rules are involved in the deviations of a particular text or group of texts. It follows that it is possible to describe a grammar of deviations from the rule and thereby to parameterize the level to which an NLP system can be made sensitive and led to a controlled acceptance of such deviations.

Also, as the syntactic structures are described as a list of pattern frames, it is possible at any time to easily insert new structures, also deviating from the standard or standard use of the language. These last two characteristics allow you to calibrate the system to communication and textual types (utterances) which are very different and particular, such as the dictionary and encyclopedia definitions (often characterized by autonomous nominal phrases or infinitive phrases), the particular prose of administrative texts and also the apparently informal and ungrammatical language of social networks.

**3. Conclusions**

This article aimed at a first presentation of SYNTAGMA, a grammar driven parsing system. In contrast to statistical and data driven systems, the system's behavior may be better controlled and fine-tuned. Its theoretical background, moving from Tesnière's valency grammar (1959) and integrating notions coming also from other frameworks, suggests some solutions for managing well known problems in Dependency Parsing (coordination, gaps, long-term relations), which have been briefly discussed here. The system is structured as a two-component grammar: the first one defines general grammar rules which are language independent, and the second is language specific and is registered in the system's resource databases. Therefore, although the actual implementation is focused on Italian, an extension to other languages seems possible, without structural changes of the core engine and the general grammar. Further experiments will follow, in order to verify this hypothesis.

SYNTAGMA's prototype has also been tested in its adaptability to different text types: some constraints may be relaxed, some others may be strengthened, depending on the target *corpus*. Experiments have been made with dictionary entries, which show some particular features: independent infinitive clauses, autonomous NP and AdjP, coordinated sequences of PP adjuncts to coordinated NPs, which generate structural and attachment ambiguities. By integrating syntactic and semantic information given by the Semantic Net of the system, the purpose is to enhance SYNTAGMA's performance in this domain.

# 4. References

Gennaro Chierchia and Sally McConnell-Ginet. 2000. *Meaning and Grammar.* MIT Press , Cambridge, Mass.

Noam Chomsky. 1988. *Language and Problems of Knowledge. The Managua Lectures.* The MIT Press, Cambridge, Mass.

Tullio De Mauro. 1971. *Per una teoria formalizzata del noema lessicale e della storicità e socialità dei fenomeni linguistici*. In *Senso e significato. Studi di semantica teorica e storica.* Adriatica, Bari.

Charles Fillmore. 1968. *The case for case*. In Bach and Harms (editors). *Universals in Linguistic Theory,* pp. 1-88, Rinehart and Winston. Holt.

Liliane Haegeman. 1991. *Introduction to Government & Binding Theory.* Basil Blackwell, Oxford.

Ray Jackendoff. 1986. *Semantics and Cognition.* MIT Press, Cambridge, Mass.

Philip N. Johnson Laird. 1983. *Mental Models,. Towards a Cognitive Science of Language.* Cambridge University Press, Cambridge (trad. it. *Modelli mentali.* Il Mulino, Bologna 1988).

Vincenzo Lombardo and Leonardo Lesmo. 1998. *Unit Coordination and Gapping in Dependency Theory*, *Proc. Workshop on Processing of Dependency Parsing*, Montral pp.11-20.

Daniel Gildea and Martha Palmer. 2002. *The Necessity of Parsing for Predicate Argument Recognition.* In *Proceedings of the 40th Annual Meeting of the Association for Computational Linguistics*, pp. 239-246, Philadelphia.

Hans Glinz. 1994. *Grammatiken im Vergleich.* Max Niemeyer, Tübingen.

Matteo Grella, Marco Nicola and Daniel Christen. 2011. *Experiments with a Constraint-based Dependency Parser*, Evalita 2011 Dependency Parsing Task.

William Labov, 1972. *Where do Grammars Stop?* in *Report of the Tewnty-Third Annual Round Tabel Meeting on Linguistics and Language Studies*, R.W.Shuy (ed.) Georgetown Univ. Press, Washington D.C.

Ronald W. Langacker. 1995. *Structural Syntax: The View from Cognitive Grammar.* In Sémiotiques, n° 6-7, décembre 1994, pp. 69-84.

Ryan McDonald, Fernando Pereira, Kiril Ribarov and Jan Hajič. 2005. *Non-projective Dependency Parsing using Spanning Tree Algorithms.*

Umberto Eco. 1984. *Semiotica e filosofia del linguaggio*. Bompiani, Milano.

Joachim Nivre. 2005. *Dependency Grammar and Dependency Parsing*.

François Rastier. 1991. *Sémantique et recherches cognitives.* PUF, Paris.

Lorenzo Renzi *et al.*, editors, 1988, 1991, 1995. *Grande grammatica italiana di consultazione.* 3 voll., Il Mulino, Bologna.

Luigi Rizzi. 1986. *Null objects in Italian and the theory of pro*, "Linguistic Inquiry", 17, pp. 501-58.

Luigi Rizzi. 1990. *Spiegazione e teoria grammaticale.* Unipress, Padova.

Raffaele Simone. 1990. *Fondamenti di linguistica.* Laterza, Roma-Bari.

Lucien Tesnière. 1959. *Éléments de syntaxe structurale.* Editions Klincksieck, Paris.